\documentclass{article}

 \usepackage[dblblindworkshop, final]{neurips_2025}
 \usepackage{tabularx}
\usepackage{latexsym}
\usepackage{amssymb}
\usepackage{amsmath}
\usepackage{amsthm}
\usepackage{booktabs}
\usepackage{enumitem}
\usepackage{graphicx}
\usepackage{color}
\usepackage{multirow}
\usepackage{array}
\usepackage{subcaption}
\usepackage[linesnumbered,ruled,vlined]{algorithm2e}

\usepackage[utf8]{inputenc} 
\usepackage[T1]{fontenc}    
\usepackage{hyperref}       
\usepackage{url}            
\usepackage{booktabs}       
\usepackage{amsfonts}       
\usepackage{nicefrac}       
\usepackage{microtype}      
\usepackage{xcolor}         

\title{Training-Time Explainability for Multilingual Hate Speech Detection: Aligning Model Reasoning with Human Rationales}
\workshoptitle{MusIML}

\author{%
 M.D.M Qureshi\dag \thanks{Corresponding Author. \dag Student Authors} \\
 Technological University Dublin\\
 \texttt{D22124696@mytudublin.ie} \\
 \And
 Sannaan Khan\dag\\  
 National University of Sciences and Technology\\
 \texttt{mkhan.msse24sines@student.nust.edu.pk }
 \And
 M. Atif Qureshi \\
 Technological University Dublin\\
 \texttt{atif.qureshi@tudublin.ie}\\
 \And
 Wael Rashwann \\
 Maynooth University\\
 \texttt{wael.rashwan@mu.ie} 
}

\begin{document}

\maketitle
\vspace{-4mm} 
\begin{abstract}
Online hate against Muslim communities often appears in culturally coded, multilingual forms that evade conventional AI moderation. Such systems, though accurate, remain opaque and risk bias, over-censorship, or under-moderation, particularly when detached from sociocultural context. We propose a \emph{training-time} explainability framework that aligns model reasoning with human-annotated rationales, improving both classification performance and interpretability. Our approach is evaluated on HateXplain (English) and BullySent (Hinglish), reflecting the prevalence of anti-Muslim hate across both languages. Using LIME, Integrated Gradients, Grad$\times$Input, and attention, we assess accuracy, explanation quality, and cross-method agreement. Results show that gradient- and attention-based regularization improve F-scores, enhance plausibility and faithfulness, and capture culturally specific cues for detecting implicit anti-Muslim hate, offering a path toward multilingual, culturally aware content moderation.
\end{abstract}

\section{Introduction}
Online hate against Muslim communities often appears in subtle, implicit, or culturally coded forms, frequently expressed in multilingual or code-switched language such as Hinglish. These complexities challenge moderation systems, risking both \emph{under-moderation}, which allows harmful narratives to persist, and \emph{over-moderation}, which suppresses legitimate religious or cultural expression \cite{gillespie2018custodians, nobata2016abusive}. Given its prevalence in both English and South Asian online spaces, tackling such content is essential for Muslim and Global digital safety and inclusivity.

While transformer-based systems achieve high accuracy, they remain opaque and risk reinforcing biases \cite{udupa2023ethical, siapera2022ai}, especially when divorced from sociocultural context. For Muslim communities, like other underrepresented groups, these shortcomings can amplify harm and erode trust in AI moderation.

Explainable AI (XAI) offers transparency \cite{abdul2018trends}, but post-hoc methods applied after training rarely influence how models learn, limiting their ability to produce faithful, culturally aware explanations \cite{floridi2018AI4people, sambasivan2021everyone, qureshi2023toward, qureshi2025explainable}.

We address this gap with a multilingual, training-time XAI regularization framework that aligns model reasoning with human rationales. Using RoBERTa and XLM-R classifiers, and four explanation strategies - LIME \cite{ribeiro2016should}, Integrated Gradients \cite{sundararajan2017axiomatic}, Grad$\times$Input \cite{shrikumar2016not}, and attention \cite{bahdanau2014neural}-we guide models toward linguistically and culturally meaningful features. Our evaluation on HateXplain (English) and BullySent (Hinglish) demonstrates cross-linguistic generalization for detecting implicit anti-Muslim hate.

\noindent\textbf{Contributions:} (1) a training-time XAI regularization framework aligning reasoning with human rationales; (2) comparative evaluation of four integrated XAI methods; (3) extension to a multilingual, code-switched setting; (4) analysis of plausibility, faithfulness, and cross-method agreement, highlighting improved detection of implicit, culturally coded anti-Muslim hate; and (5) evidence that gradient- and attention-based regularization offer the best balance of accuracy, explanation quality, and efficiency.\footnote{Code and Data at https://github.com/DeedahwarMazhar/NeurIPS-MulsimsInML-TrainingTimeReg}

\section{Methodology}

\subsection{Datasets and Pre-processing}
We evaluate on two culturally and linguistically distinct datasets to test monolingual and multilingual performance. 

\textbf{HateXplain} \cite{mathew2021hatexplain} contains $\sim$20k English social media posts (Twitter, Gab) labeled as Hate, Offensive, or Normal, with multi-annotator token-level rationales. Following prior work, Hate and Offensive are merged into a \emph{Detrimental Content} class, including \textit{15}\% posts explicitly targeting Muslims. 

\textbf{BullySent} is a Hinglish (Hindi–English code-switched) cyberbullying dataset from Indian social media, comprising $\sim$\textit{6.5k} posts. While it does not explicitly mark anti-Muslim hate, it labels religion-based hate, serving as a relevant proxy for South Asian contexts where Hinglish is the dominant medium of informal online discourse \cite{bhange2020hinglishnlp}. For comparability, all labels are unified into overarching categories: \emph{Detrimental} (e.g., Hate, Offense, Bullying) vs. \emph{Non-Detrimental} content.

\subsection{Model Architectures}
We employ two transformer encoders to match the linguistic characteristics of each dataset: RoBERTa-base for HateXplain (English) and XLM-RoBERTa-base for BullySent (Hinglish). Each encoder is followed by a Dense ReLU layer, a dropout of 0.3, and a sigmoid output. Models are optimized using Adam with a learning rate of $2 \times 10^{-5}$.

\subsection{Training-Time Explanation-Based Regularization}
Our training augments binary cross-entropy loss with an explanation alignment term, applied only when the gold label is \emph{Detrimental Content}.

The classification loss is:
\begin{equation}
    Loss_{BCE} = -\frac{1}{N}\sum^{N}_{i=0}y_{t}\log(y_{p})-(1-y_{t})\log(1-y_{p})
\end{equation}
where $y_t$ is the true label and $y_p$ the predicted probability.

Let $rat(i)$ and $exp(i)$ denote normalized rationale and explanation scores for each term $i$ in the text input. We aggregate explanation score at word level to ensure compatibility with human-annotated rationales. The explanation loss is:
\begin{equation}
Loss_{EXP}(rat \parallel exp) = \sum_{i} rat(i) \log \left( \frac{rat(i)}{exp(i)} \right)
\end{equation}

The total loss is:
\begin{equation}
    Loss_{COMP} = Loss_{BCE} + \lambda  Loss_{EXP}(rat \parallel exp)
\end{equation}
with $\lambda \in \{0.001, 0.005, 0.01, 0.05, 0.1, 0.5, 1, 5, 10\}$ tuned by validation F1-score and early stopping.

Algorithm~\ref{alg:training} summarizes the process. We use the non-regularized model as our baseline for comparison.

\begin{algorithm}[t]
\small
\caption{Training with Explanation-Based Regularization}
\label{alg:training}
\KwIn{Training set $D = \{(x_i, y_i, rat_i)\}_{i=1}^N$, XAI method $E$, regularization weight $\lambda$}
\KwOut{Trained model $f_\theta$}

\For{each epoch}{
    \For{each batch $(x, y, rat)$ in $D$}{
        $y_p \gets f_\theta(x)$ \tcp*{Forward pass}
        $Loss_{BCE} \gets \text{BinaryCrossEntropy}(y, y_p)$
        
        \If{  $y == 1$}{ 
        $exp \gets E(f_\theta, x)$ \tcp*{Generate explanation}
        \If{$exp$ is token-level}{
                $exp \gets \text{ConvertTokenToWord}(exp)$ \tcp*{Aggregate word scores}
            }
        $Loss_{EXP} \gets \text{KL}(rat \parallel exp)$ \tcp*{Explanation loss}
        
        }

        $Loss_{COMP} \gets Loss_{BCE} + \lambda \cdot Loss_{EXP}$

        $\theta \gets \theta - \eta \cdot \nabla_\theta Loss_{COMP}$ \tcp*{Backpropagation}
    }
    \If{early stopping criterion met on validation set}{
        \textbf{break}
    }
}
\end{algorithm}

\subsection{XAI Methods}
We compare four complementary explanation strategies, covering both model-agnostic and model-specific paradigms:

\begin{itemize}
    \item \textbf{LIME} \cite{ribeiro2016should}: Model-agnostic and perturbation-based; fits a sparse linear surrogate by masking tokens. We limit $n\_samples$ to 128 for efficiency.
    \item \textbf{Integrated Gradients (IG)} \cite{sundararajan2017axiomatic}: Integrates gradients from a zero embedding baseline to the input over 15 steps, capturing non-linear dependencies.
    \item \textbf{Grad$\times$Input (GXI)} \cite{shrikumar2016not}: Multiplies token embeddings with their gradients to estimate local feature sensitivity in one backward pass.
    \item \textbf{Attention} \cite{bahdanau2014neural}: Uses averaged final-layer attention weights across heads as token importance; not always faithful but internally consistent for regularization.
\end{itemize}

\subsection{Post-Hoc Explanation Evaluation}
We evaluate XAI methods by comparing post-hoc attributions to human-annotated rationales. Scores are normalized to probability distributions over tokens, and subword attributions (IG, GXI, Attention) are merged to word-level before alignment. Word scores are binarized into rationale masks using:
\begin{enumerate}
    \item \textbf{Top-$K$:} highest $K \in \{3,5,7,10,15,20,25\}$ tokens (LIME: $K \leq 10$).
    \item \textbf{Thresholding:} tokens with scores $>\theta \in \{0.001,0.005,0.01,0.05,0.1,0.5\}$.
\end{enumerate}
Human rationales are aggregated via the \emph{Union} strategy.

We report:
\begin{itemize}
    \item \textbf{Plausibility:} \emph{IoU} (overlap proportion) and \emph{Token-F1} (token-level precision–recall) vs. human rationales.
    \item \textbf{Faithfulness:} \emph{Prediction Drop}—confidence decrease when identified tokens are removed.
\end{itemize}

This token-level alignment captures fine-grained linguistic/contextual cues, and comparing Top-$K$ vs. thresholding highlights trade-offs between fixed- and variable-length rationales.

\section{Results and Discussion}

\begin{table*}[t]
\centering
\small
\setlength{\tabcolsep}{4pt}
\renewcommand{\arraystretch}{1.15}
\begin{tabular}{|l|l|c|c|c|c|c|c|c|c|}
\hline
\multirow{2}{*}{\textbf{Method}} & \multirow{2}{*}{\textbf{Data}} & \multicolumn{2}{c|}{\textbf{Perf.}} & \multicolumn{2}{c|}{\textbf{IoU}} & \multicolumn{2}{c|}{\textbf{Token-F1}} & \multicolumn{2}{c|}{\textbf{Drop}} \\ \cline{3-10}
 &  & Acc & F1 & Top-K & Thresh. & Top-K & Thresh. & Top-K & Thresh. \\ \hline
\multirow{2}{*}{No Regularizer} 
  & HX & 0.78 & 0.82 & — & — & — & — & — & — \\ \cline{2-10}
  & BS & 0.84& 0.83& — & — & — & — & — & — \\ \hline
\multirow{2}{*}{LIME} 
  & HX & 0.80 & 0.84 & 0.09 & 0.11 & 0.19 & 0.22 & 0.25 & 0.26 \\ \cline{2-10}
  & BS & 0.85& 0.84& \textbf{0.27}& 0.11& \textbf{0.45}& \textbf{0.34}& 0.41& 0.45\\ \hline
\multirow{2}{*}{Integrated Gradients} 
  & HX & 0.79 & 0.83 & 0.21 & 0.28 & 0.39 & 0.44 & \textbf{0.72} & \textbf{0.74} \\ \cline{2-10}
  & BS & 0.84& 0.84& 0.14& 0.12& 0.23& 0.22& 0.41& \textbf{0.82}\\ \hline
\multirow{2}{*}{Grad$\times$Input} 
  & HX & 0.78 & 0.82 & 0.21 & 0.19 & 0.38 & 0.35 & 0.63 & 0.35 \\ \cline{2-10}
  & BS & 0.85& 0.85& 0.07& 0.06& 0.16& \textbf{0.12}& \textbf{0.68}& 0.54\\ \hline
\multirow{2}{*}{Attention} 
  & HX & \textbf{0.80} & \textbf{0.84} & \textbf{0.33} & \textbf{0.38} & \textbf{0.58} & \textbf{0.60} & 0.64 & 0.60 \\ \cline{2-10}
  & BS & \textbf{0.86}& \textbf{0.85}& 0.11& 0.12& 0.21& 0.17& 0.30& 0.30\\ \hline
\end{tabular}
\caption{Combined classification and XAI performance. HX = HateXplain, BS = BullySent. Perf. = Accuracy, F1. IoU and Token-F1 measure plausibility; Drop measures faithfulness. Best scores per metric (HX and BS) are in bold.}
\label{tab:bigresults}
\end{table*}

Table~\ref{tab:bigresults} reports classification (Acc, F1), plausibility (IoU, Token-F1), and faithfulness (Prediction Drop) for HateXplain (HX) and BullySent (BS).

\textbf{Classification.} Attention achieves the best overall F1 (HX: \textbf{0.84}, BS:\textbf{ 0.85}) and highest BS accuracy (\textbf{0.86}). LIME and IG perform competitively on HX, while G$\times$I is efficient but slightly less accurate.

\textbf{Plausibility.} On HX, Attention leads in IoU/Token-F1 (\textbf{0.38}/\textbf{0.60}), with IG second. On BS, LIME attains the best Top-K plausibility (IoU \textbf{0.27}, Token-F1 \textbf{0.45}), suggesting perturbations capture Hinglish cues better than attention or gradients.

\textbf{Faithfulness.} IG yields the largest drops on both datasets (HX: \textbf{0.72}/\textbf{0.74}; BS: \textbf{0.82}), indicating strong causal linkage. G$\times$I performs well on BS, while LIME’s drops remain low. Gradient-based methods transfer faithfulness more robustly across languages, while plausibility varies by dataset and binarization strategy.

\begin{figure*}[t]
    \centering
    \begin{subfigure}[t]{0.3\textwidth}
        \centering
        \includegraphics[width=\textwidth]{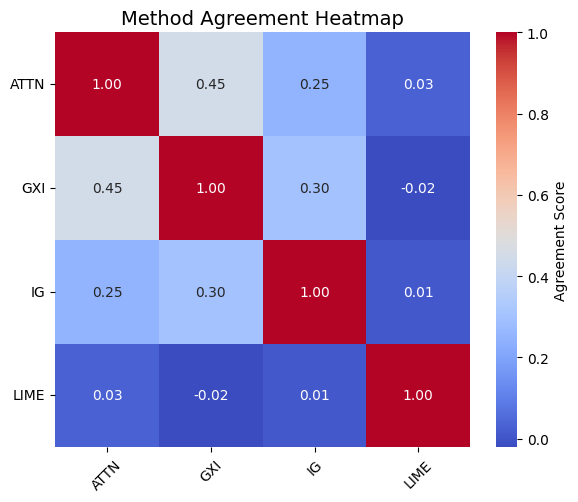}
        \caption{Method Agreement Heatmap for XAI methods with Kendall's Agreement Correlation ($\tau$).}
        \label{fig:qual_a}
    \end{subfigure}
    \hfill
    \begin{subfigure}[t]{0.69\textwidth}
        \centering
        \includegraphics[width=\textwidth]{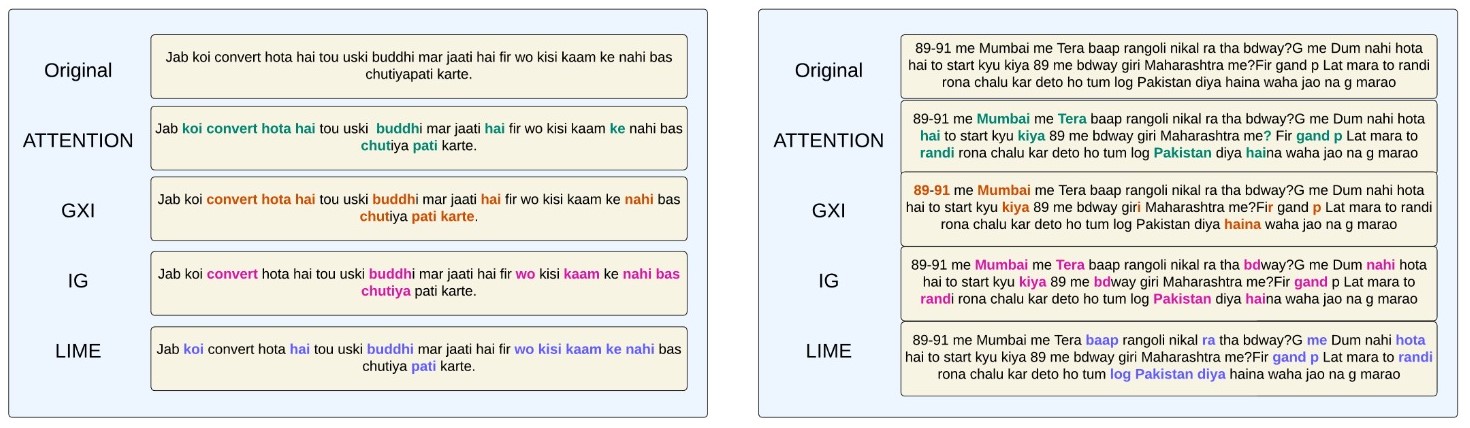}
        \caption{Qualitative comparison of token-level attributions from four XAI methods across two code-switched hate speech examples. While gradient-based methods (Attention, G$\times$I, IG) frequently assign salience to subword fragments, LIME operates at the word level and often misses parts of compound slurs.}

        \label{fig:qual_b}
    \end{subfigure}
    \hfill

   \caption{Kendall’s $\tau$ agreement between XAI methods (left), showing higher correlation among gradient-based approaches and divergence from LIME. Qualitative token-level attributions for two representative code-switched hate speech examples (right) illustrate method-specific highlighting patterns.}

    \label{fig:qualitative}
\end{figure*}

\textbf{Inter-Method Agreement:} We measure similarity via Kendall’s $\tau$, suitable for comparing ordered token attributions. Figure~\ref{fig:qual_a} shows highest alignment between Attention and G$\times$I ($\approx 0.45$), with moderate scores for Attention–IG and G$\times$I–IG, and near-zero or negative correlations for LIME due to its word-level attributions.

\textbf{Qualitative Comparison.} Figure~\ref{fig:qual_b} visualizes token-level highlights for a representative post with different XAI methods. Gradient-based approaches often mark subword units within compounds, enabling finer-grained detection of culturally specific or offensive expressions. In contrast, perturbation-based methods tend to operate at the whole-word level and can overlook embedded cues.

\section{Conclusion}
We present a training-time explainability framework that embeds human-aligned rationales into hate speech classifiers, improving both plausibility and faithfulness across English and Hinglish datasets. Results show that gradient- and attention-based regularization generalize well across languages, capturing culturally specific cues often missed by perturbation-based methods, and offering a path toward more accountable multilingual moderation.
\section*{Acknowledgemets}
This publication has emanated from research conducted with the financial support of Research Ireland (RI) Center for Research Training in Machine Learning (ML-Labs), under Grant number 18/CRT/6183. For Open Access, the author has applied a CC BY public copyright license to any Author Accepted Manuscript version arising from this submission.

\bibliographystyle{unsrt}  
\bibliography{ref}

\appendix

\section{Related Work}

AI-based content moderation has evolved from keyword matching and simple classifiers \cite{davidson2017automated} to transformer-based models such as BERT \cite{devlin2019bert} and multilingual encoders \cite{conneau2020unsupervised}. While these improve accuracy, they still struggle with sarcasm, implicit hate \cite{vidgen2020directions}, and culturally coded language, often producing opaque or biased decisions across linguistic and cultural boundaries \cite{waseem2021disembodied}. Human-in-the-loop approaches \cite{kiela2021dynabench} highlight that moderation requires cultural and political context \cite{udupa2023ethical, siapera2022ai}, especially for communities facing targeted hostility.

Explainable AI (XAI) offers interpretability via methods such as LIME \cite{ribeiro2016should}, SHAP \cite{lundberg2017unified}, attention \cite{bahdanau2014neural}, and counterfactuals \cite{sen2022counterfactually, qian2021counterfactual}. However, post-hoc explanations can be unfaithful \cite{jain2019attention, serrano2019attention} and often lack integration with domain knowledge, which is critical for detecting subtle, culturally specific hate speech against marginalized groups.

Training-time XAI regularization, enabled by datasets like HateXplain \cite{mathew2021hatexplain}, aligns model explanations with human rationales \cite{ross2017right, rieger2020interpretations, carton2021learn}, but most work uses a single XAI method and monolingual data. Few evaluate both plausibility and faithfulness \cite{jacovi2020towards, wiegreffe2021measuring} or assess consistency across explanation methods. This gap is pronounced in multilingual and code-switched contexts like Hinglish, where implicit anti-Muslim hate is prevalent. Our work addresses this by integrating multiple XAI methods into training, applying them to both English and Hinglish datasets, and jointly evaluating cultural relevance, explanation quality, and cross-method agreement (Table \ref{tab:literature}).

\begin{table}[t]
\renewcommand{\arraystretch}{1.5}
\centering
\begin{tabular}{|m{3.5cm}|m{2cm}|m{2.5cm}|m{2cm}|m{2cm}|}
\hline
\textbf{Related Studies} & \textbf{Multiple XAI Methods} & \textbf{XAI-based Regularization} & \textbf{Explanation Optimization} & \textbf{XAI Evaluations}\\
\hline
Mathew et al. \cite{mathew2021hatexplain} & & \checkmark & & \checkmark\\\hline
Ross et al. \cite{ross2017right} & \checkmark & \checkmark & & \\\hline
Hancock et al. \cite{hancock2018training} & & \checkmark & \checkmark & \\\hline
Balkir et al. \cite{balkir2022necessity} & \checkmark & & & \checkmark \\\hline
\textbf{Our Study} & \checkmark & \checkmark & \checkmark & \checkmark \\\hline
\end{tabular}
\vspace{1em}
\caption{Comparison of selected related studies with our work.}
\label{tab:literature}
\end{table}

\section{Extended Methodology}
\label{appendix:methodology}

\subsection{Datasets}
We evaluate on two datasets with token-level human rationales:  
\textbf{HateXplain} \cite{mathew2021hatexplain} contains $\sim$19k social media posts from Twitter and Gab, annotated for \textit{Hate}, \textit{Offensive}, or \textit{Normal} content, along with rationales from multiple annotators. Following prior work, we merge \textit{Hate} and \textit{Offensive} into a single \textbf{Detrimental Content} label.  

\textbf{BullySent} is a Hinglish (Hindi–English code-switched) dataset of $\sim$6.4k posts annotated for abusive content, also with token-level rationales. Compared to HateXplain, BullySent contains greater lexical variation, frequent spelling inconsistencies, and transliteration noise, making rationale alignment more challenging.  

For evaluation, we aggregate annotator rationales using the \textbf{Union} strategy to ensure inclusive coverage of all highlighted tokens.

\begin{table}[t]
\renewcommand{\arraystretch}{1.25}
\centering
\begin{tabular}{|m{1.25cm}|m{1.5cm}|m{1.5cm}|m{1cm}|}
\hline
\textbf{Dataset} & \textbf{Detrimental} & \textbf{Non-Detrimental} & \textbf{Total} \\
\hline
HateXplain & 11,415 & 7,814 & 19,299 \\
BullySent  & 3,451  & 2,985 & 6,436  \\
\hline
\end{tabular}
\caption{Dataset distribution across both corpora.}
\label{tab:dataset_distribution}
\end{table}

\subsection{Training Pipeline}
\begin{figure}[t]
    \centering
    \includegraphics[width=0.85\linewidth]{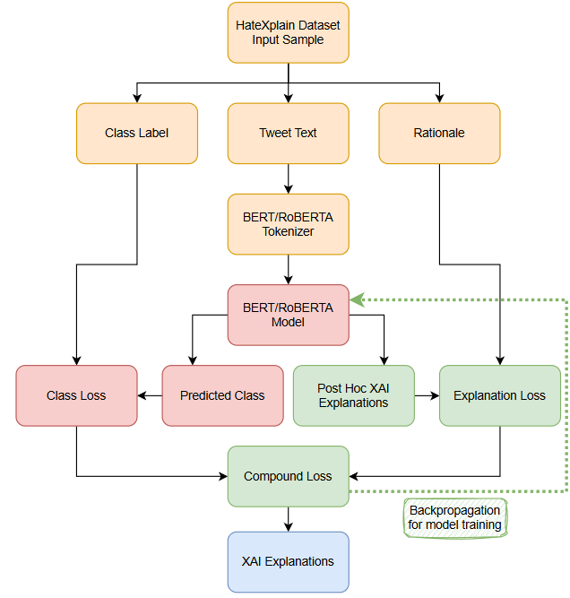}
    \caption{Training pipeline with integrated explanation regularization.}
    \label{fig:overview}
\end{figure}

Figure~\ref{fig:overview} outlines our training procedure. For HateXplain, we fine-tune a RoBERTa encoder, while for BullySent (Hinglish), we use XLM-R to better handle multilingual and code-switched text. Both models employ a binary classification head and are optimized with a composite loss combining standard binary cross-entropy with an explanation-alignment term. For correctly predicted detrimental-content examples, we generate token-level importance scores using a given XAI method, normalize them, and align them with human rationales via KL divergence.  

The regularization weight $\lambda$ controls the trade-off between accuracy and explanation alignment, tuned separately for each dataset. The optimal value was $\lambda = 5$ for HateXplain and $\lambda = 0.05$ for BullySent, suggesting that in the multilingual, noisier BullySent setting, heavy regularization can hurt performance, whereas HateXplain benefits from stronger alignment pressure.  

This approach is applied consistently across all four XAI methods studied: Attention, Gradient$\times$Input, Integrated Gradients, and LIME.

\section{Extended Explainability Analysis Across Datasets}

\begin{figure}[t]
    \centering
    \includegraphics[width=0.32\textwidth]{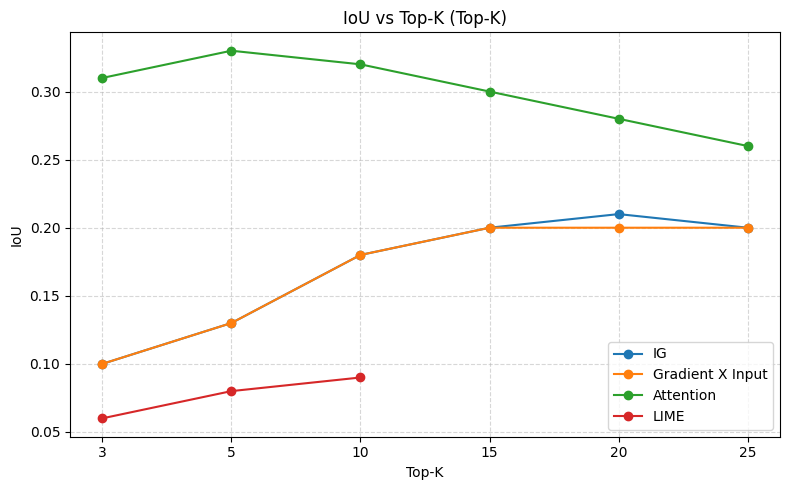}
    \includegraphics[width=0.32\textwidth]{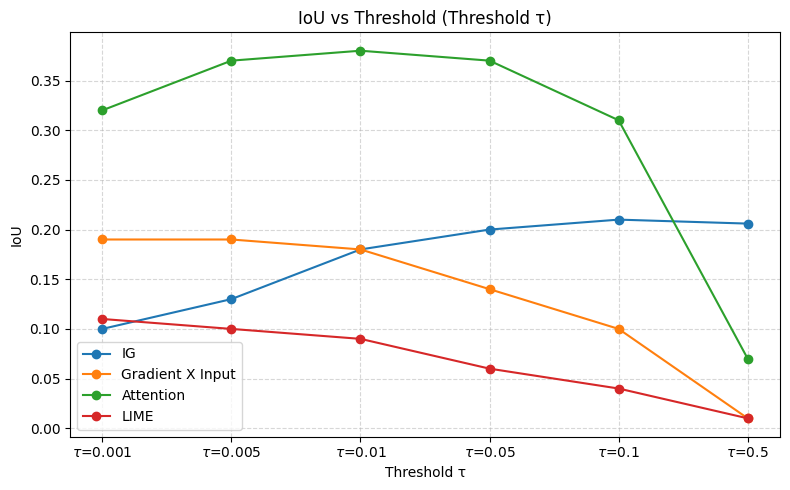}
    \includegraphics[width=0.32\textwidth]{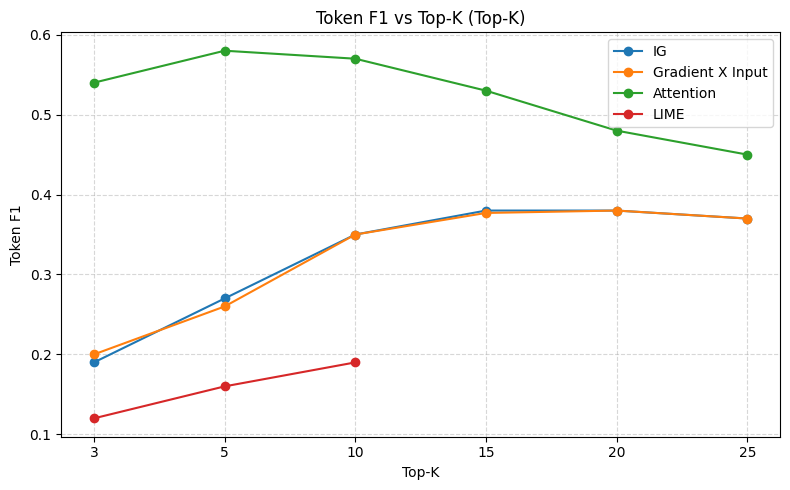} \\
    \includegraphics[width=0.32\textwidth]{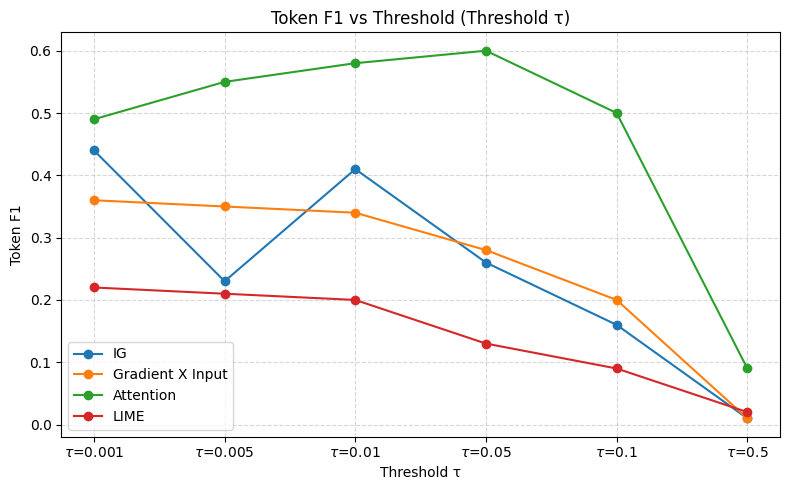}
    \includegraphics[width=0.32\textwidth]{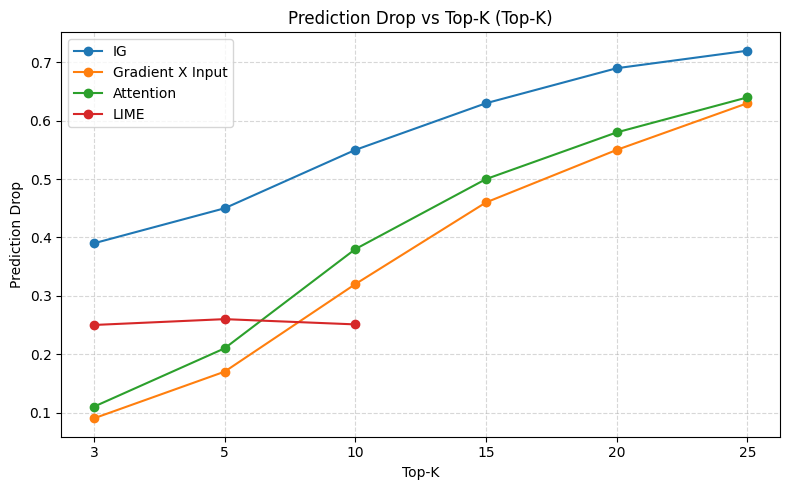}
    \includegraphics[width=0.32\textwidth]{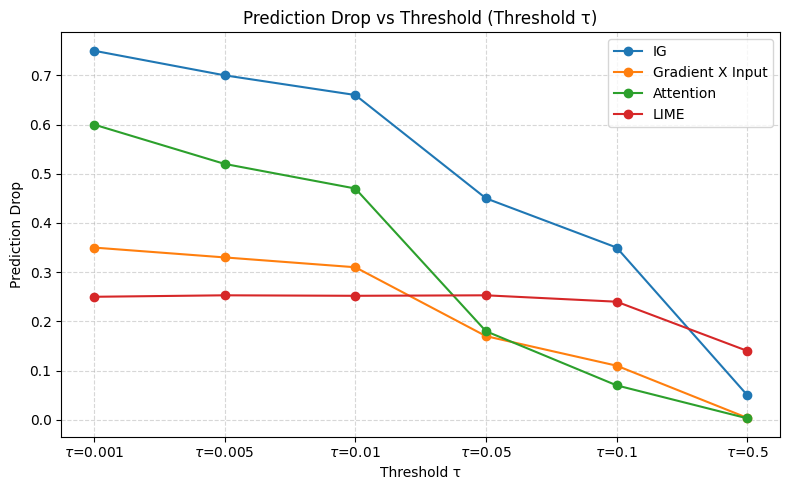} \\
    \includegraphics[width=0.32\textwidth]{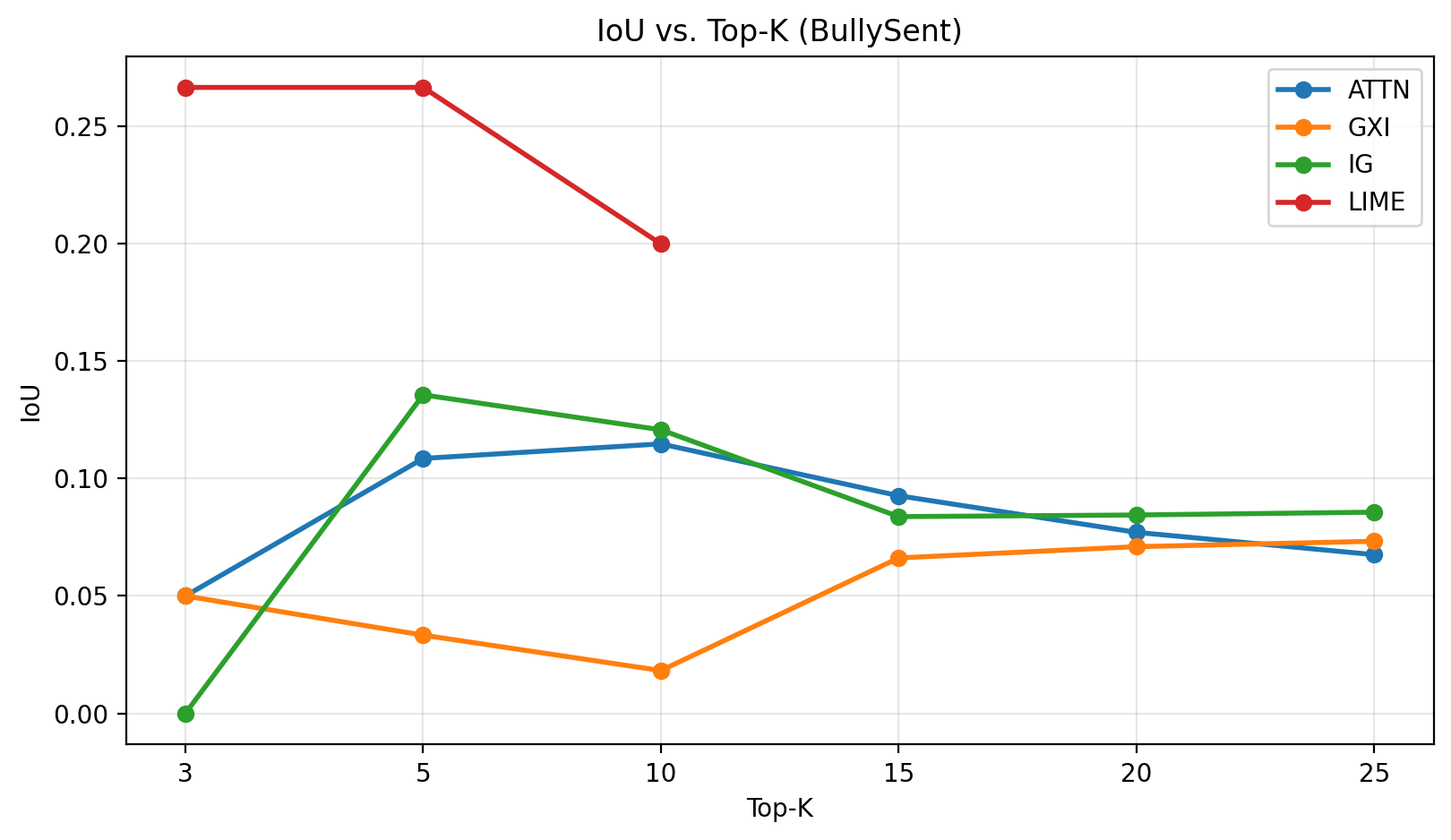}
    \includegraphics[width=0.32\textwidth]{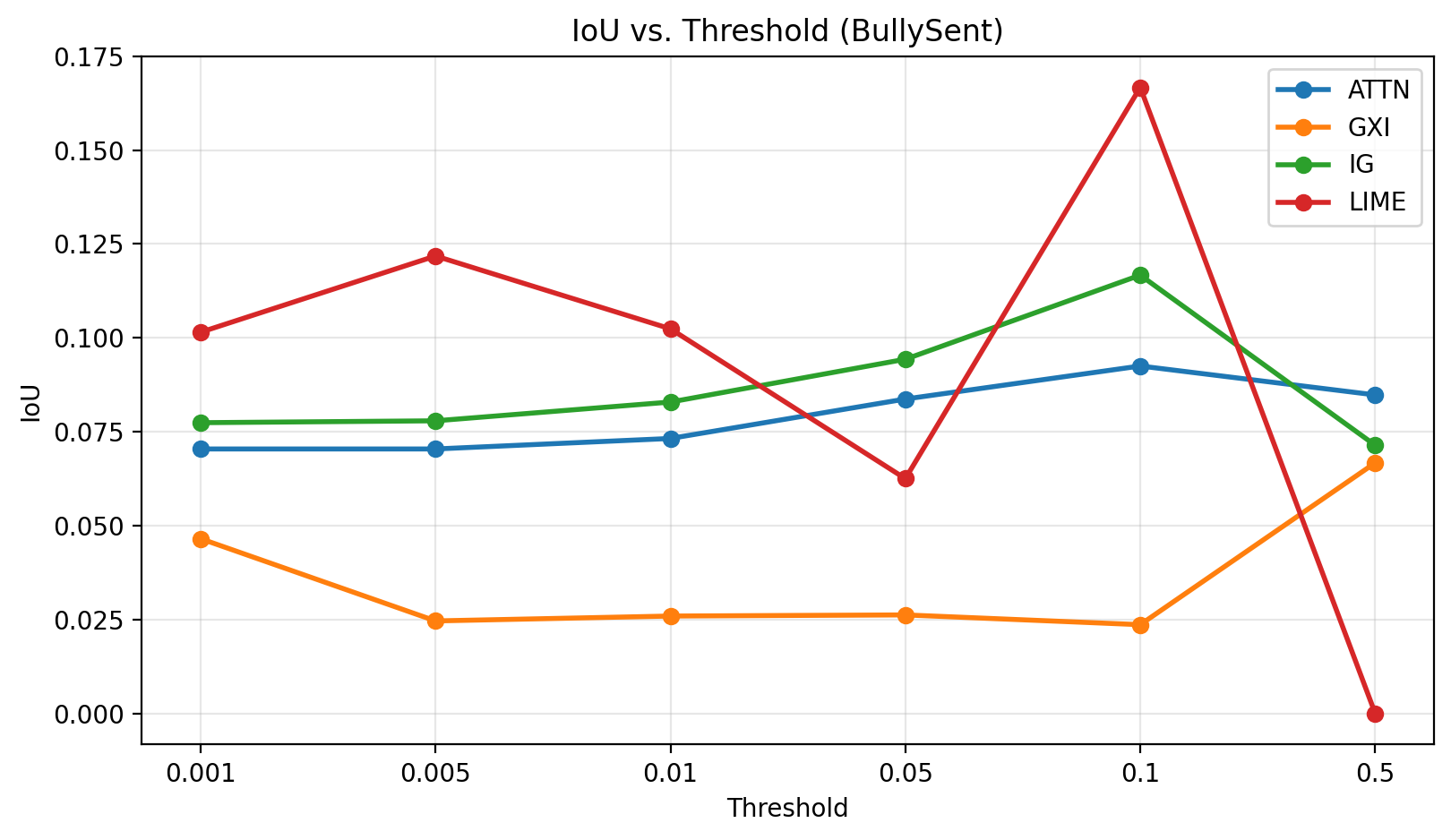}
    \includegraphics[width=0.32\textwidth]{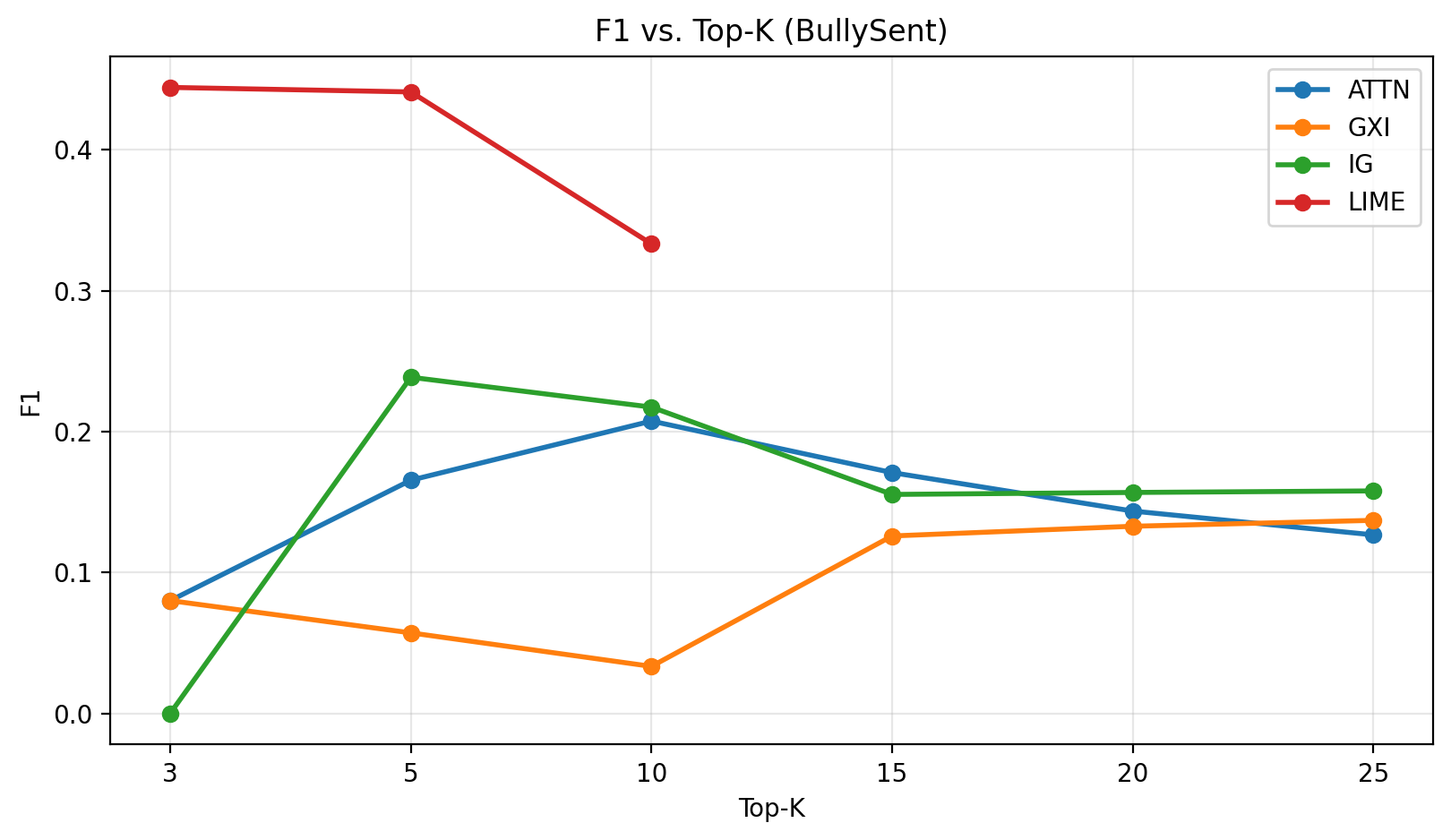} \\
    \includegraphics[width=0.32\textwidth]{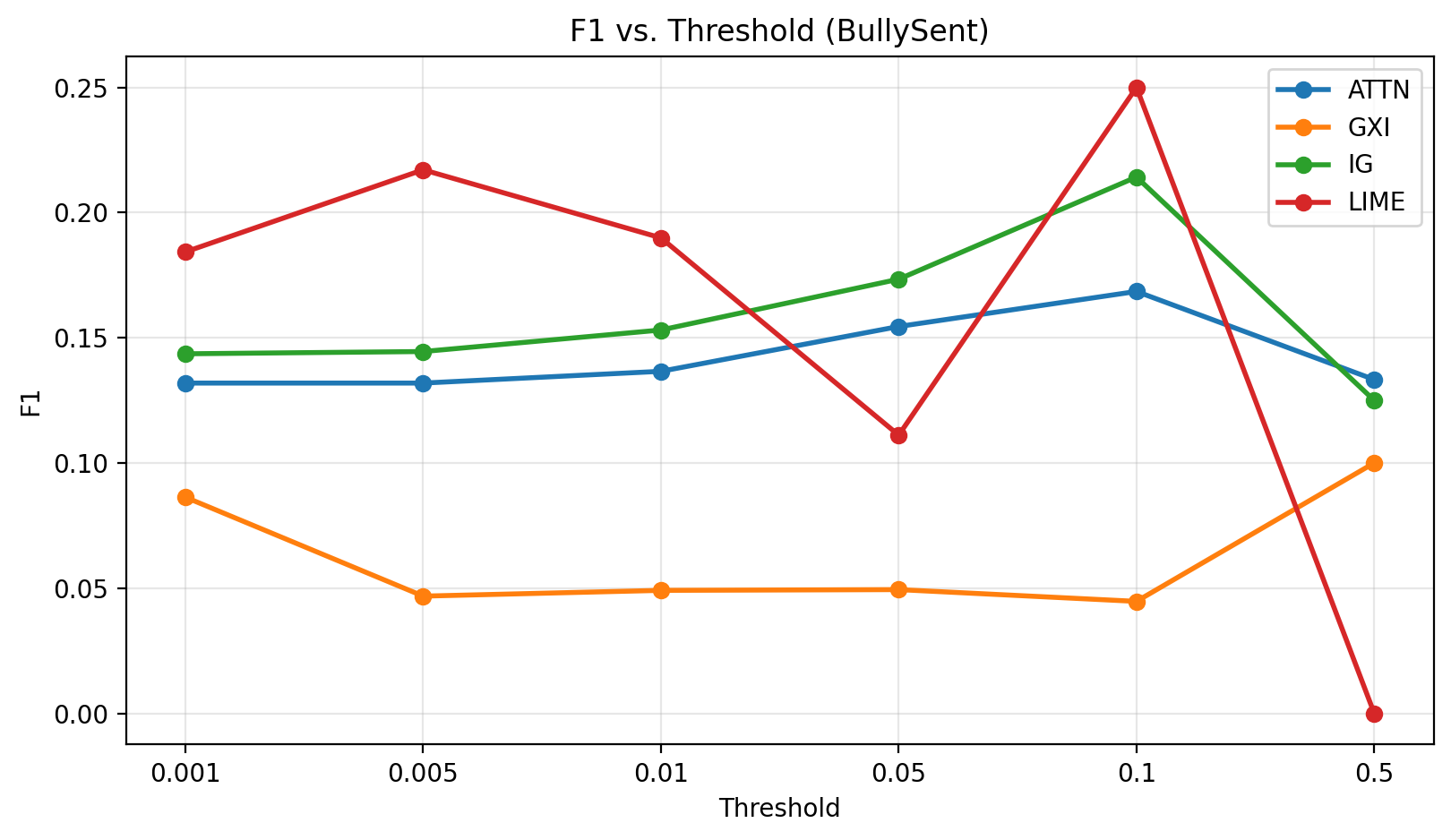}
    \includegraphics[width=0.32\textwidth]{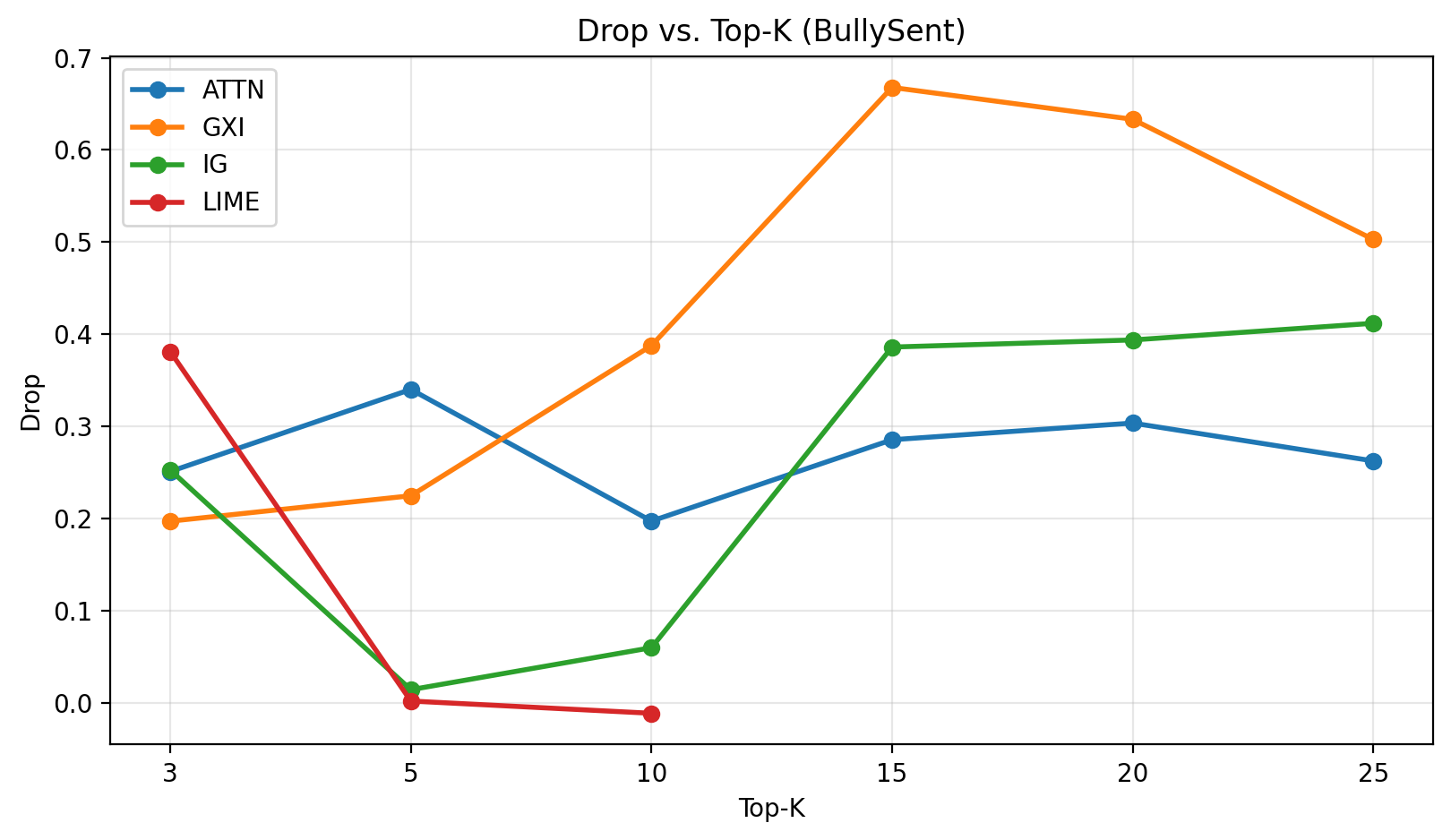}
    \includegraphics[width=0.32\textwidth]{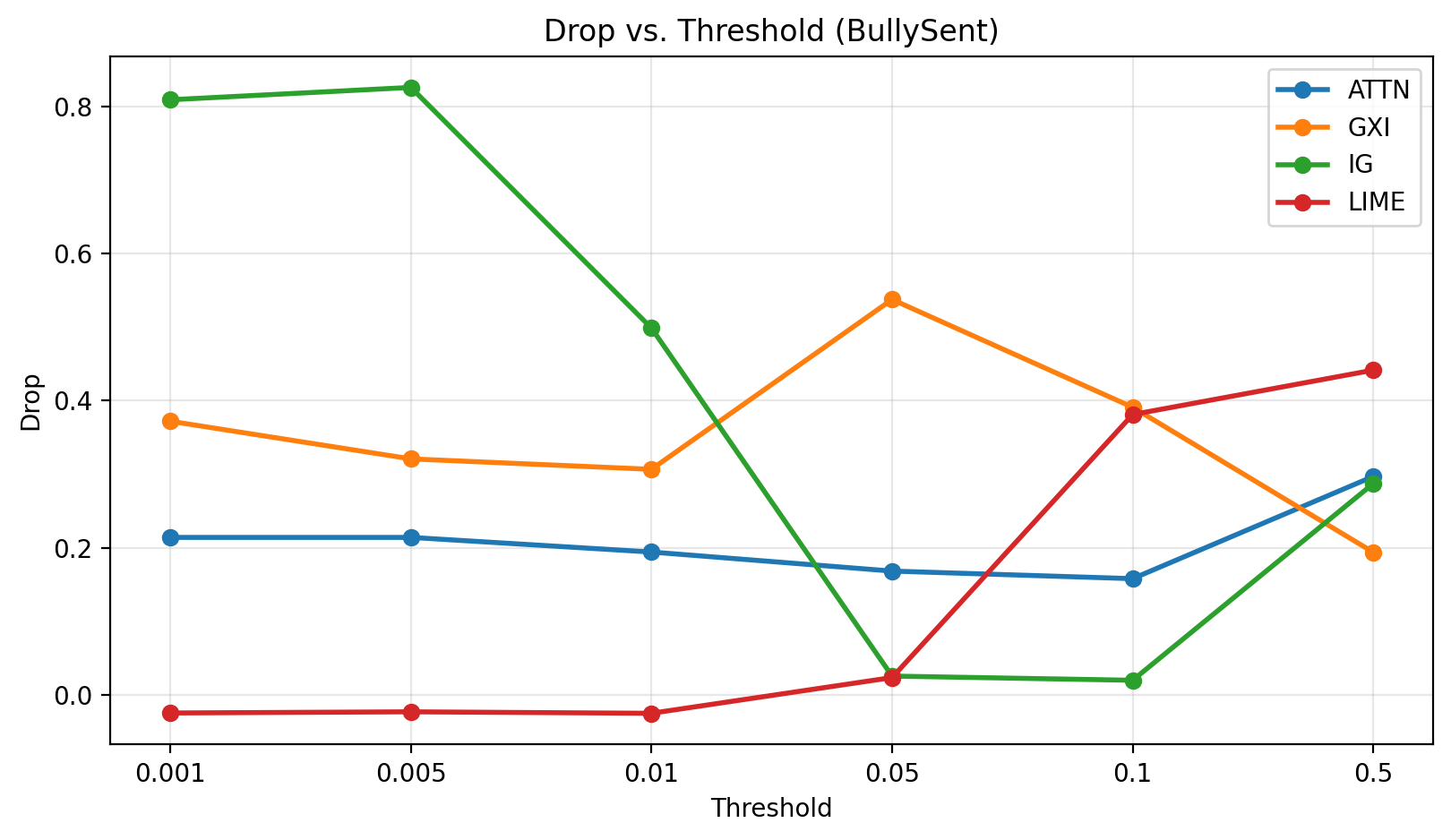}
    \caption{Variation of IoU, Token-F1, and Prediction Drop across Top-K and Threshold values for HateXplain (top two rows) and BullySent (bottom two rows). LIME’s Top-K curves terminate at $K{=}10$ due to sampling constraints.}
    \label{fig:combined_xai_plots}
\end{figure}

To assess the generality of our findings, we evaluate explanation plausibility and faithfulness on both HateXplain and BullySent using the same four XAI methods (Attention, Gradient$\times$Input, Integrated Gradients, and LIME) and two rationale selection strategies (Top-K and Thresholding). For each dataset, we plot metric trends over varying $K$ and $\tau$ values, with all curves using identical scales to enable direct visual comparison. Figures~\ref{fig:combined_xai_plots} present all twelve plots (six per dataset), while the following discussion synthesizes the results across datasets.

\subsection{Intersection over Union}
Across both datasets, Attention consistently achieves the highest IoU values under both selection strategies, indicating strong alignment with human rationales. Integrated Gradients (IG) follows closely, especially under thresholding, where it benefits from capturing non-linear token dependencies. Gradient$\times$Input lags in plausibility on both datasets, while LIME exhibits the lowest IoU scores, particularly for higher $K$ values. Notably, BullySent scores are generally lower than HateXplain across methods, suggesting that shorter, noisier, and more code-switched inputs make exact token overlap harder to achieve.

\subsection{Token-F1}
Token-level F1 patterns are remarkably consistent across datasets: attention explanations best match individual human-annotated tokens, followed by IG. Thresholding continues to outperform Top-K for most methods, reflecting that selecting tokens above a score threshold better preserves rationale boundaries—especially beneficial for HateXplain’s longer texts. On BullySent, all methods show reduced F1 scores relative to HateXplain, likely due to higher annotation sparsity and linguistic variability, but the relative ranking of methods remains stable.

\subsection{Prediction Drop}
Prediction Drop trends reaffirm IG as the most faithful method in both datasets: masking its top-ranked tokens produces the largest confidence decreases, indicating strong causal alignment with the model’s decision process. Attention also yields substantial drops, though with more variability across $K$ and $\tau$. LIME maintains low prediction drop values across datasets, reinforcing critiques of its faithfulness in high-dimensional, context-dependent NLP settings. Interestingly, BullySent exhibits smaller overall drop values compared to HateXplain, which may be due to its shorter, more direct statements where a few key tokens suffice for classification.

\end{document}